\documentclass[a4paper,twoside]{article}

\usepackage{epsfig}
\usepackage{subcaption}
\usepackage{calc}
\usepackage{amssymb}
\usepackage{amstext}
\usepackage{amsmath}
\usepackage{amsthm}
\usepackage{multicol}
\usepackage{pslatex}
\usepackage{apalike}
\usepackage[bottom]{footmisc}

\usepackage{amsfonts,bm} 
\usepackage{algorithm}
\usepackage{algorithmic}
\usepackage{graphicx}
\usepackage{bibunits}  
\usepackage{booktabs}
\usepackage{makecell}
\usepackage{color}
\usepackage{url}
\usepackage[T1]{fontenc}

\usepackage{SCITEPRESS}     

\theoremstyle{plain}

\newcommand{\RR}{\mathbb{R}}                                     

\newcommand{\bi}{\begin{itemize}}
\newcommand{\ei}{\end{itemize}}
\newcommand{\ba}{\begin{array}}
\newcommand{\ea}{\end{array}}

\DeclareMathOperator*{\argmin}{arg\,min}
\DeclareMathOperator*{\argmax}{arg\,max}

\newcommand{\pivec}{\vec{\pi}}

\newcommand{\avec}{\vec{a}}
\newcommand{\E}{\mathbb{E}}

\newif\ifarxiv
\arxivtrue

\begin{document}

\title{Adaptive action supervision in reinforcement learning \\from real-world multi-agent demonstrations}

\ifarxiv
\author{\authorname{Keisuke Fujii\sup{1,2,3}\orcidAuthor{0000-0001-5487-4297}, Kazushi Tsutsui\sup{1}, Atom Scott\sup{1}, \\ Hiroshi Nakahara\sup{1}, Naoya Takeishi\sup{4,2}, Yoshinobu Kawahara\sup{5,2}}
\affiliation{\sup{1}Graduate School of Informatics, Nagoya University, Nagoya, Aichi, Japan.}
\affiliation{\sup{2}Center for Advanced Intelligence Project, RIKEN, Osaka, Osaka, Japan.}
\affiliation{\sup{3}PRESTO, Japan Science and Technology Agency, Tokyo, Japan}
\affiliation{\sup{4}The Graduate School of Engineering, The University of Tokyo, Tokyo, Japan}
\affiliation{\sup{5}Graduate School of Information Science and Technology, Osaka University, Osaka, Japan}
\email{$^a$ fujii@i.nagoya-u.ac.jp
}
}
\else
\author{Anonymous}
\fi
\vspace{-10pt}
\vspace{-10pt}
\abstract{Modeling of real-world biological multi-agents is a fundamental problem in various scientific and engineering fields. Reinforcement learning (RL) is a powerful framework to generate flexible and diverse behaviors in cyberspace; however, when modeling real-world biological multi-agents, there is a domain gap between behaviors in the source (i.e., real-world data) and the target (i.e., cyberspace for RL), and the source environment parameters are usually unknown. In this paper, we propose a method for adaptive action supervision in RL from real-world demonstrations in multi-agent scenarios. We adopt an approach that combines RL and supervised learning by selecting actions of demonstrations in RL based on the minimum distance of dynamic time warping for utilizing the information of the unknown source dynamics. This approach can be easily applied to many existing neural network architectures and provide us with an RL model balanced between reproducibility as imitation and generalization ability to obtain rewards in cyberspace. In the experiments, using chase-and-escape and football tasks with the different dynamics between the unknown source and target environments, we show that our approach achieved a balance between the reproducibility and the generalization ability compared with the baselines. In particular, we used the tracking data of professional football players as expert demonstrations in football and show successful performances despite the larger gap between behaviors in the source and target environments than the chase-and-escape task.}
\vspace{-10pt}
\maketitle 

\vspace{-10pt}
\section{\uppercase{Introduction}}
\label{sec:introduction}
\vspace{-5pt}
Modeling real-world biological multi-agents is a fundamental problem in various scientific and engineering fields.
For example, animals, vehicles, pedestrians, and athletes observe others' states and execute their own actions in complex situations.
Pioneering works have proposed rule-based modeling approaches such as in human pedestrians \cite{Helbing95} and animal groups \cite{Couzin02} for each domain using hand-crafted functions (e.g., social forces). 
Recent advances in reinforcement learning (RL) with neural network approaches have enabled flexible and diverse modeling of such behaviors often in cyberspace \cite{ross2010efficient,ho2016generative}.

\begin{figure*}[ht]
\centering
\includegraphics[width=0.75\linewidth]{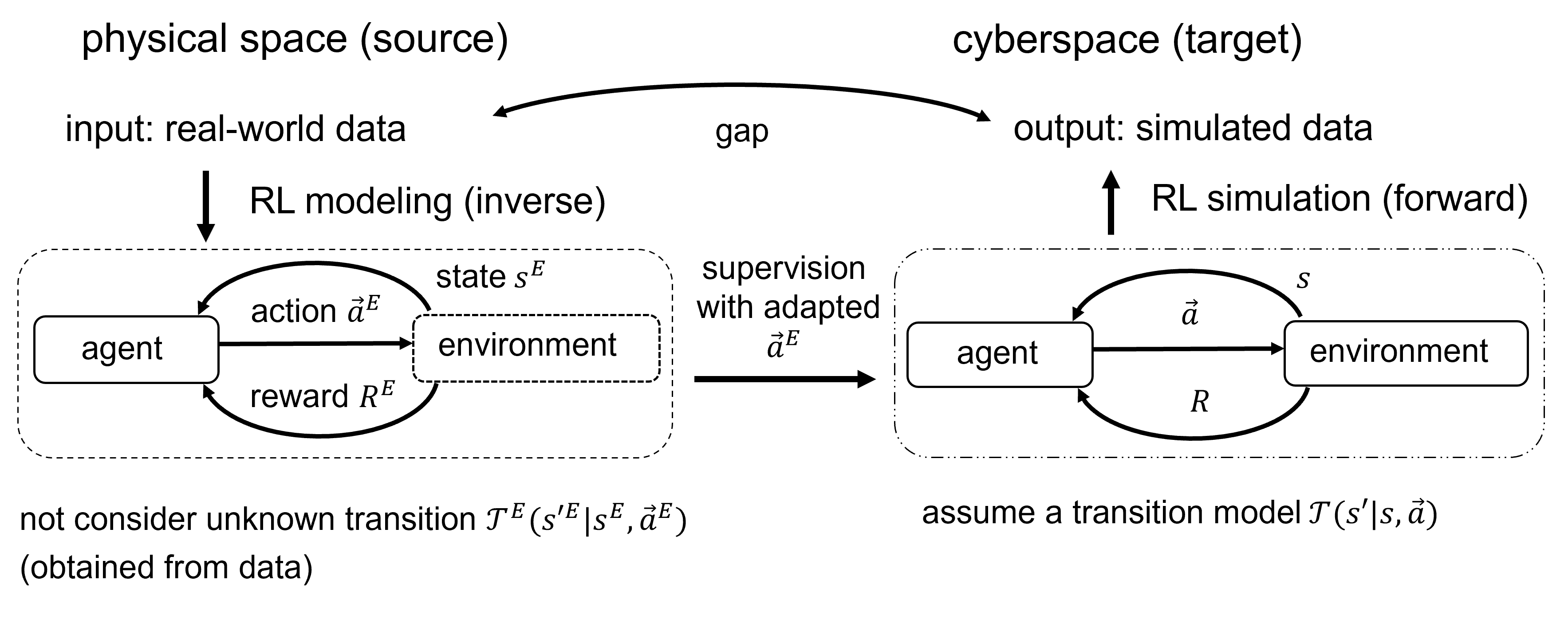}
\vspace{-10pt}
\caption[]{Our problem setting and solution. We consider a Real-to-Sim domain adaptation problem in which the source and target are real-world data in a physical space and simulated data in cyberspace, respectively. 
We first perform inverse RL modeling from real-world data, but we do not consider unknown transition model $\mathcal{T}^E$ because we can obtain the next state from demonstration data. 
We then perform forward RL with a temporal transition model $\mathcal{T}$, but there should be a discrepancy between $\mathcal{T}^E$ and $\mathcal{T}$. 
Since we cannot access the information of $\mathcal{T}^E$, our solution comprises the supervised learning using the observed joint action $\avec$ and adapted $\avec^E$ in cyberspace for the learning of the Q function (see also Section \ref{sec:proposed}). 
}
\vspace{-10pt}
\label{fig:diagram}
\end{figure*}

However, when modeling real-world biological multi-agents, domain gaps may occur between behaviors in the sources (real-world data) with unknown dynamics and targets (cyberspace in RL) as shown in Fig. \ref{fig:diagram}. 
The opposite configuration of the source and target has been actively studied and known as Sim-to-Real \cite{rusu2017sim}, which transfers the knowledge from cyberspace or human demonstrations to almost known source dynamics (simulation in Sim-to-Real) such as real-world robotics \cite{schaal1996learning,kolter2007hierarchical}.
In contrast, domain adaptation in real-world situations where the parameters of the source environment are often unknown cannot utilize explicit dynamics regarding source environments (e.g. transition model). 
In other words, we consider a Real-to-Sim domain adaptation problem in which the unknown source and the target are real-world data in a physical space and simulated data in cyberspace, respectively (Fig. \ref{fig:diagram}). 
Team sports (e.g., football) and biological multi-agent motions (e.g., chase-and-escape) are examples that can be addressed with the above approach.
Agents observe others' states  and execute planned actions \cite{fujii2021data}, and particularly in team sports, most of the governing equations are unknown.
In such complex real-world behaviors, studies have separately investigated RL in cyberspace for flexible adaptation to complex environments \cite{kurach2020google,li2021celebrating} and data-driven modeling for reproducing real-world behaviors \cite{Zheng16,Le17,fujii2021learning}. 
However, given the gap between these forward and backward approaches (see Fig. \ref{fig:diagram}) in multi-agent RL (MARL) scenarios, an integrated approach to combine both strengths will be required.

In this paper, we propose a method for adaptive action supervision in RL from real-world multi-agent demonstrations.
To utilize the information of unknown source dynamics, we adopt an approach that combines RL and supervised learning by selecting the action of demonstrations based on the minimum distance between trajectories in source and target environments.
Our goal is to balance reproducibility as imitations and generalization ability to obtain rewards (e.g., when different initial values are given in the same environment).
These are mutually independent in general, and our approach that combines RL and supervised learning will help us achieve our goal. 
Compared with the case of Sim-to-Real in robotics, there have been no simulators for real-world agents with explicit dynamics in biological multi-agent scenarios.
In our experiments, we use simple chase-and-escape and football tasks with different dynamics between unknown source and target environments.

In summary, our main contributions are as follows.
(1) We propose a novel method for adaptive action supervision in RL from multi-agent demonstration, which will bridge the gap between RL in cyberspace and real-world data-driven modeling. 
(2) We adopt an approach that combines RL and supervised learning by selecting actions of demonstrations in RL based on the minimum distance of dynamic time warping (DTW) \cite{vintsyuk1968speech} to utilize the information of the unknown source dynamics. This approach can be easily applied to existing neural network architectures and provide an RL model balanced between reproducibility as imitation and generalization ability. 
(3) In the experiments, using a chase-and-escape and football tasks with the different dynamics between the unknown source and target environments, our approach struck a balance between the reproducibility and generalization compared with the baselines. In particular, we used the tracking data of professional football players as expert demonstrations in a football RL environment and show successful performances.
Our framework can estimate values for real agent behaviors and decision making if the model can imitate behaviors of players, which may be difficult for either data-driven and RL approaches. 

In the remainder, we describe the background of our problem and our method in Sections \ref{sec:background} and \ref{sec:proposed}, overview related works in Section \ref{sec:related}, and present results and conclusions in Sections \ref{sec:experiments} and \ref{sec:conclusion}.


\vspace{-15pt}
\section{\uppercase{Background}}
\label{sec:background}
\vspace{-5pt}
Here, we consider a sequential decision-making setting of multiple agents interacting with a fully observable environment.
We consider a forward RL model in Fig. \ref{fig:diagram} right, defined as a tuple $(K, S, A, \mathcal{T}, R, \gamma)$, 
where $K$ is the fixed number of agents; 
$S$ is the set of states $s$;
$A=[A_1, ..., A_K]$ represents the set of joint action $\avec \in A$ (for a variable number of agents), and $A_k$ is the set of local action $a_k$ that agent $k$ can take;
$\mathcal{T}(s'| s,\avec): S \times A \times S \rightarrow [0,1]$ is the transition model for all agents;
$R=[R_1, ..., R_K]: S \times A \rightarrow \RR^K$ is the joint reward function;
and $\gamma \in (0,1]$ is the discount factor. %
In on-policy RL, the agent learns a policy $\pi_{k}: S_k \times A \rightarrow [0,1]$, where $S_k$ is a set of states for $k$.

The objective of agent $k$ is to discover the policy $\pi_k$ that maps states to actions, thereby maximizing the expected total reward over the agent's lifespan, i.e., $G_k = \sum_{t}^{T} \gamma^{t} R_{k,t}$, where $R_{k,t}$ is the reward of agent $k$ at time $t$ and $T$ is the time horizon. 
The value $Q^\pi_k(s_k,a_k)$ related to a specific state-action pair $(s_k,a_k)$ serves as a expected  future reward that can be acquired from $(s_k,a_k)$ when adhering to policy $\pi_k$. The optimal value function $Q^*(s,a)$, offering the maximal values across all states, is determined by the Bellman equation:
\vspace{-5pt}
{\small
\begin{equation}
Q_k^*(s_k,a_k) = \E\left[{R_k(s_k,a_k) + \gamma \sum_{s_k'} \mathcal{T}_k(s_k'|s_k,a_k) \max_{a_k'} Q_k^*(s_k',a_k')}\right],
\label{eq:bellman}
\end{equation}
}

\noindent where $\mathcal{T}_k$ is the transition model of agent $k$.
The optimal policy $\pi_k$ is then $\pi_k(s_k) = \argmax_{a_k\in A}Q_k^*(s_k,a_k).$
Since our approach can be easily applied to existing neural network in model-free RL, we consider both independent policy for each agent and the joint policy $\pivec$ inducing the joint action-value function $Q_{tot}^{\pivec}(s, \avec)=\mathbb{E}_{s_{0:\infty},\avec_{0:\infty}}\left[\sum_{t=0}^\infty \gamma^t R_t \mid s_0=s, \avec_0=\avec, \pivec \right]$, where $R_t$ is the value of the joint reward at time $t$.

In a multi-agent system in complex real-world environments (e.g., team sports), (i) transition functions are difficult to design explicitly. 
Instead, (ii) if we can utilize the demonstrations of expert behaviors (e.g., trajectories of professional sports players), we can formulate and solve it as a machine learning problem (e.g., learning from demonstration).
In other words, if the problem falls in the case that satisfies the two conditions, (i) and (ii), learning from demonstration is a better option than a pure RL approach by constructing the environment without demonstrations.
As shown in Fig. \ref{fig:diagram} right, we perform forward RL with a temporal transition model $\mathcal{T}(s'|s,\avec)$, but there should be a discrepancy between $\mathcal{T}$ and $\mathcal{T}^E(s'^E|s^E,\avec^E)$ in Fig.\ref{fig:diagram} left. 
Since we cannot access the information of $\mathcal{T}^E$, our solution is the supervised learning using the observed joint action $\avec$ and adapted $\avec^E$ in cyberspace for the learning of the Q function (see also Section \ref{sec:proposed}).
Next, we introduce DQN framework according to the previous work \cite{hester2018deep}.
For simplicity, we describe the following explanation using a single-agent RL framework and omit the agent index $k$.

DQN leverages a deep neural network to approximate the value function $Q(s,a)$ \cite{mnih2015human}. 
The network is designed to generate a set action values $Q(s,\cdot;\theta)$ for a given state input $s$, where $\theta$ represents the network's parameters. 
DQN employs a separate target network, which is duplicated from the main network after every $\tau$ steps to ensure more consistent target Q-values.
The agent records all of its experiences in a replay buffer $\mathcal{D}^{replay}$, which is subsequently uniformly sampled for network updates.

The double Q-learning updates the current network by computing the argmax over the subsequent state values and uses the target network for action value \cite{van2016deep}. 
The loss for Double DQN (DDQN) is defined as:
{\small
\begin{equation}
    J_{DQ}(Q) = \sum^{T-1}_t \left( R_t + \gamma Q(s_{t+1}, a^{\max}_{t+1}; \theta') - Q(s_t, a_t; \theta) \right)^2,   
\label{eq:ddqn}
\end{equation}
}
where $\theta'$ refers to the parameters of the target network, and $a^{\max}_{t+1} = \argmax_{a_t} Q(s_{t+1}, a_t; \theta)$. 
The upward bias typically associated with regular Q-learning updates is reduced by separating the value functions employed for these two variables.
For more efficient learning, e.g., to sample more significant transitions more frequently from its replay buffer, prioritized experience replay~\cite{schaul2016prioritized} have been used.

\section{\uppercase{Adaptive action supervision in RL}}
\label{sec:proposed}
\vspace{-5pt}
In many real-world settings of RL, we can access observation data of the multi-agent system, but we cannot access an accurate model of the system. 
To construct an alternative simulator, we want the agent to learn as much as possible from the demonstration data. 
In particular, we aim to decrease the domain gap between behaviors in the source data and the target environments. 
Here, we describe our adaptive action supervision approach for RL from demonstrations.
We adopt the following three steps according to the deep Q-learning from demonstrations (DQfD) \cite{hester2018deep}.
The first is pre-training, which learns to imitate the demonstrator.
The second is sampling actions from the pre-trained RL model in the target RL environment.
The third is training in the RL environment. 
During the pre-training and training phases, the network is updated with mainly two losses: the 1-step double Q-learning loss in Eq. (\ref{eq:ddqn}) and a dynamic time-warping supervised classification loss. 
As mentioned above, the Q-learning loss ensures that the network satisfies the Bellman equation and can be used as a starting point for TD learning.
For the second loss, we propose a simple supervised loss for actions and a dynamic time-warping action assignment for efficient pre-training and RL. 

The supervised loss is crucial for pre-training because the demonstration data usually covers a narrow part of the state space and does not take all possible actions.
Here we consider a single agent case for simplicity (i.e., we removed the notation of agent $k$, but we can easily extend it to multi-agent cases).
The previous DQfD \cite{hester2018deep} introduces a large margin classification loss~\cite{piot2014boosted}: 
{\small
\begin{equation}
J_{MS}(Q) = \sum^T_t \max_{a_t \in A}[Q(s_t, a_t) + l(a^E_t, a_t)] - Q(s_t, a^E_t),
\label{eq:marginclassification}
\end{equation}
}
where $a^E_t$ is the action the expert demonstrator takes in state $s_t$ and $l(a^E_t, a_t)$ is a margin function that is $0$ when $a_t = a^E_t$ and positive otherwise. 
This loss makes the value of the expert's action higher than the other actions' values, at least by the margin $l$.
This approach would be effective for learning maximum Q-function values; however, when limited data are available, the direct approach to maximize the Q-function values for the action of the demonstration may be efficient. 
Therefore, we propose a simple supervised loss for actions represented by the cross-entropy of softmax values of the Q-function such that
\vspace{-5pt}
\begin{equation}
J_{AS}(Q_t) = -\sum^T_t \mathbf{a}^E_{t}\cdot 
\log
\left(\mathrm{softmax}(\mathbf{q}_{s_t})
\right),
\label{eq:simplesupervised}
\end{equation}
where $\mathbf{a}_t^E \in \{0, 1\}^{|A|}$ (i.e., one-hot vector of expert actions), $\mathbf{q}_{s_t} = [ Q(s_t,a_t=1), ..., Q(s_t,a=|A|) ]$, and the $\log$ applies element-wise. 
Ideally, Q-functions in the source and target domains should be compared, but when using limited data, it would be better that more reliable action data is used as supervised data (rather than using approximated Q-function from data). 
This loss aims to achieve both reproducibility and generalization by maximizing the Q-function values for the action of the demonstration.
A similar idea has been used \cite{hester2018deep,Lakshminarayanan2016}, which used only similar supervised losses in pre-training or RL, respectively, but we explicitly define and use this loss for both pre-training and RL to balance reproducibility and generalization.

Eq. (\ref{eq:simplesupervised}) and the large margin classification loss in DQfD ~\cite{piot2014boosted} in Eq. (\ref{eq:marginclassification}) assume that the timestamp of expert action $a^E_t$ should be the same as that of the RL model $a_t$.
However, when there is a discrepancy between the source and target environments, the appropriate timestamp of expert actions can vary from that of the RL model actions.
Thus, we propose a dynamic time-warping supervised loss for actions, which does not require prior knowledge, utilizing DTW framework \cite{vintsyuk1968speech}, a well-known algorithm in many domains \cite{sakoe1978dynamic,myers1980performance,tappert1990state}. 

Here, we first consider two state sequences in RL and demonstration: $s = s_1,\ldots,s_t,\ldots,s_n$ and $s^E = s^E_1,\ldots,s^E_j,\ldots,s^E_m$, where $n$ and $m$ are the lengths of the sequences.
We select $a^E_{t'}$ at $t'$ (which is not necessarily equal to $t$) for demonstration defined as:
\begin{equation}
t' = \argmin_j W(s, s^E)_{t,j},
\label{eq:warping}
\end{equation} 
where $W(s, s^E) \in \RR^{n \times m}$ is a warping path matrix based on a local distance matrix $d(s, s^E) \in \RR^{n \times m}$ (e.g., Euclidean distance) and some constraints such as monotonicity, continuity, and boundary \cite{sakoe1978dynamic}. 
$W(s, s^E)_{t,j} \in \RR$ is the $(t,j)$ component of $W(s, s^E)$.  
Then we obtain the supervised action loss with adaptive action supervision by modifying Eq. (\ref{eq:simplesupervised}) such that
\vspace{-5pt}
\begin{equation}
J_{AS+DA}(Q_t) = -\sum^T_t \mathbf{a}^E_{t'}\cdot\log\left(
\mathrm{softmax}(\mathbf{q}_{s_t} )
\right).
\label{eq:adaptation}
\end{equation}
Additionally, we introduce an $\mathcal{L}_2$ regularization loss that targets the weights and biases of the network, aiming to avoid overfitting given the small size of the demonstration dataset. 
The total loss used to update the network is as follows:
\begin{equation}
  J(Q) = J_{DQ}(Q) + {\lambda_1}J_{AS+DA}(Q) + {\lambda_2}J_{\mathcal{L}_2}(Q).
  \label{eq:losses}
\end{equation}
The $\lambda$ parameters control the weight of these losses.
As an ablation study, we examine removing some of these losses in Section~\ref{sec:experiments}.
The behavior policy is $\epsilon$-greedy based on the Q-values.
Note that, similarly to DQfD \cite{hester2018deep}, after the pre-training phase is finished, the agent starts interacting with the environment, collecting its own data, and adding it to its replay buffer  $\mathcal{D}^{replay}$. 
The agent overwrites the buffer when the buffer is full, but does not overwrite the demonstration data.

\vspace{-20pt}
\section{\uppercase{Related Work}}
\label{sec:related}
\vspace{-5pt}
In RL from demonstration \cite{schaal1996learning}, the direct approach recovers experts’ policies from demonstrations by supervised learning \cite{pomerleau1991efficient,ross2010efficient,Ross11} 
or generative adversarial learning \cite{ho2016generative,song2018multi}, which make the learned policies similar to the expert policies (reviewed e.g., by \cite{ramirez2022model} and \cite{da2019survey,zhu2020transfer} as transfer learning). 
However, it is sometimes challenging to collect high-quality (e.g., optimal) demonstrations in many tasks. 
To obtain better policies from demonstrations, several methods combine imitation learning and RL such as  \cite{silver2016mastering,hu2018knowledge,Lakshminarayanan2016}. 
Some approaches \cite{vecerik2017leveraging,hester2018deep} have been proposed to explore the sparse-reward environment by learning from demonstrations. 
Recently, cross domain adaptation problems have been considered to achieve the desired movements such as when changing morphologies \cite{raychaudhuri2021cross,fickinger2022cross}. 
In this case, imitation in terms of reproducibility would be difficult in principle because the problem (e.g., morphology) is changed. 
Our problem setting is different in terms of achieving both the ability to maximize a reward and reproducibility (imitation ability) rather than only the former. 

These methods are often designed for single-agent tasks and attempt to find better policies by exploring demonstration actions. 
Many MARL algorithms have been proposed by modifying single-agent RL algorithms for a multi-agent environment. One of the early approaches is independent learning where an agent learns its own policy independently of the other agents \cite{omidshafiei2017deep,tampuu2017multiagent}.
Recently, in learning from demonstrations, for example, researchers have proposed MARL as a rehearsal for decentralized planning
\cite{kraemer2016multi}, MARL augmented by mixing demonstrations from a centralized policy
\cite{lee2019improved} with sub-optimal demonstrations, and centralized learning and decentralized execution \cite{peng2021hybrid}. 
Other researchers have proposed imitation learning from observations under transition model disparity \cite{gangwani2021imitation} between the dynamics of the expert and the learner by changing different configuration parameters in cyberspace.
However, domain adaptation in RL to cyberspace from real-world multi-agent demonstrations has been rarely investigated. 

In RL applications, grid-world, robot Soccer, video games, and robotics have been intensively investigated. 
Among these domains, robotics and robot soccer are specifically related to real-world problems. 
In robotics, noise in sensors and actuators, limited computational resources, and the harmfulness of random exploration to people or animals are some of the many challenges \cite{hua2021learning}.
There have been successful applications of transfer learning in robotics \cite{schaal1996learning,kolter2007hierarchical,sakato2014learning} (reviewed by e.g., \cite{zhu2021survey}).
These are mostly transferred from cyberspace or human demonstrations to real-world robotics (sometimes called Sim-to-Real \cite{rusu2017sim}), which utilize almost known dynamics about the (at least) target dynamics.
In contrast, our Real-to-Sim problem cannot utilize explicit dynamics about both source and target environments, and thus such domain adaptation in RL is challenging.

Robot soccer is similar to our task, in particular, RoboCup (the Robot World Cup Initiative) involves attempts by robot teams to actually play a soccer game \cite{Kitano97}.
Some researchers have adopted imitation learning approaches \cite{hussein2018deep,nguyen2020structure}, but the source and target environments are basically the same. 
In terms of simulators based on real-world data for data analysis, to our knowledge, there have been no domain adaptation methods in RL from real-world data to simulation environments.  

In the tactical behaviors of team sports, agents select an action that follows a policy (or strategy) in a state, receives a reward from the environment and others, and updates the state \cite{fujii2021data}. 
Due to the difficulty in modeling the entire framework from data for various reasons \cite{van2021learning}, we can adopt two approaches: to estimate the related variables and functions from data (i.e., inverse approach) as a sub-problem, and to build a model (e.g., RL) to generate data in cyberspace (i.e., forward approach, e.g., \cite{kurach2020google,li2021celebrating}). 

For the former, there have been many studies on inverse approaches.
There have been many studies on estimating reward functions by inverse RL \cite{Luo2020,Rahimian2020} and the state-action value function (Q-function) \cite{Liu2018,Liu2020,ding2022deep,nakahara2023action}. 
Researchers have performed trajectory prediction in terms of the policy function estimation, as imitation learning \cite{Le17,Teranishi2020,fujii2020policy} and behavioral modeling \cite{Zheng16,Zhan19,Yeh2019,fujii2022estimating,teranishi2022evaluation} to mimic (not optimize) the policy using neural network approaches. 
This approach did not consider the reward in RL (and simulation) and usually performed a trajectory prediction. 

For the latter approach, researchers have proposed new MARL algorithms with efficient learning, computation, and communication \cite{roy2020promoting,espeholt2019seed,liu2021semantic,li2021celebrating}. 
Recently, the ball-passing behaviors in artificial agents of Google Research Football (GFootball) \cite{kurach2020google} and professional football players were compared \cite{scott2022does}, but a gap still exists between these forward and backward approaches. 
In other research fields, e.g., for animal behavioral analysis, forward \cite{banino2018vector,ishiwaka2022deepfoids} and backward approaches \cite{ashwood2022dynamic,fujii2021learning} have also been used separately. 
Our approach integrates both approaches to combine the reproducibility as imitation and generalization to obtain rewards. 

\begin{figure*}[t]
\centering
\includegraphics[width=1\linewidth]{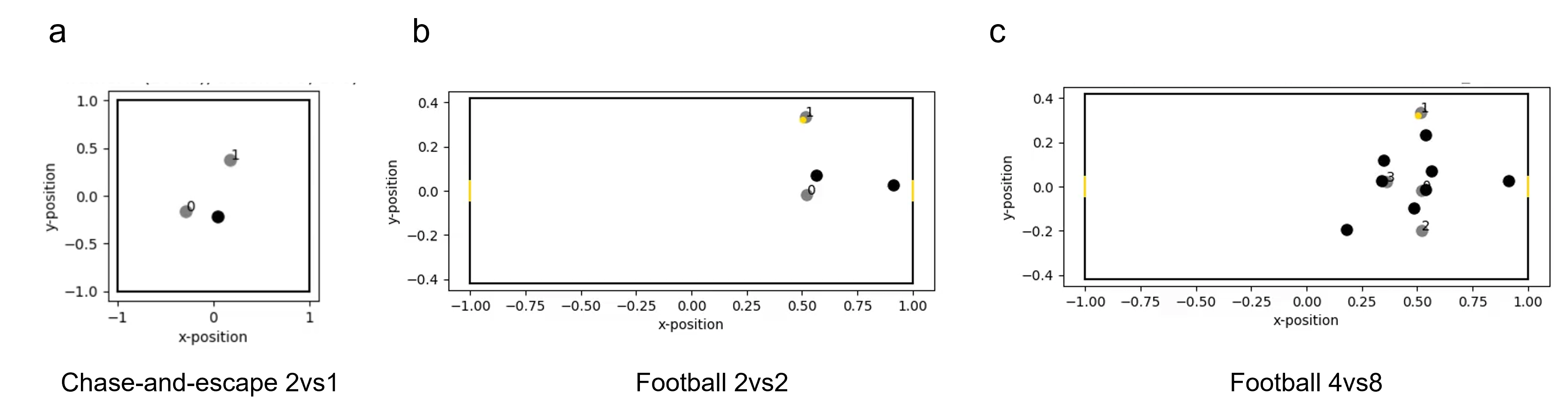}
\vspace{-10pt}
\caption[]{Our RL problem setting. (a) the predators and prey are represented as gray and black disks, respectively. 
(b) and (c) are football 2vs2 and 4vs8 tasks, respectively. Yellow circle, yellow line, gray and black circles are ball, goal line, attackers, and defenders, respectively. 
The play areas are represented by a black square/rectangle surrounding them.
}
\vspace{-10pt}
\label{fig:example}
\end{figure*}

\vspace{-15pt}
\section{\uppercase{Experiments}}
\label{sec:experiments}
\vspace{-10pt}
The purpose of our experiments is to validate the proposed methods for application to real-world multi-agent modeling,
which usually has no explicit equations in a source environment.
Hence, for verification of our methods, we first examined a simple but different simulation environment from the demonstration:
a predator-prey cooperative and competitive interaction, namely a chase-and-escape task.  
Next, we investigated a football environment with the demonstrations of real-world football players.
We considered a 2vs2 task for a simple extension of the 2vs1 chase-and-escape task and then a 4vs8 task (4 attackers) for more realistic situations. 
We basically considered decentralized multi-agent models, which do not communicate with each other (i.e., without central control) for simplicity, but in the football 4vs8 task, we also examined centralized models.  
\ifarxiv
\else
The code is provided at \url{https://drive.google.com/file/d/1H6tJ8d9JcpQL07gfBSibps_APVQ7SpEn/view?usp=sharing}.
\fi

Here, we commonly compared our full model DQAAS (deep q-learning with adaptive action supervision) to four baseline methods: a simple DQN with DDQN and prioritized experience replay introduced in Section \ref{sec:background} (without demonstration),
DQfD \cite{hester2018deep}, DQAS (deep q-learning with action supervision), and DQfAD (DQfD with adaptive demonstration using DTW). 
Using these baselines with the same network architectures for fair comparisons, we investigated the effect of adaptive action supervision. 
Note that our approach was also used in the pre-trained phase in all tasks.
We hypothesized that our approach would find a balance between imitation reproducibility and generalization compared to the baselines. 
In addition, only for the football 4vs8 task requiring more agent interaction, we examined CDS \cite{li2021celebrating}, which is a recent centralized MARL method in GFootball, as a base model.
That is, by replacing it with the above DQN, we also investigated the effectiveness of our approach as a centralized MARL method. 
Our evaluation metrics in the test phase were twofold: one is the DTW distance between the RL model and demonstration trajectories representing imitation reproducibility, and another is the obtained reward by RL agents.
We used well-known DTW distance here because it would be easy to verify whether the learning of our model is successful or not.
During the test phase, $\epsilon$ in $\epsilon$-greedy exploration was set to 0 and each agent was made to take greedy actions.
With 5 different random seeds, we evaluated the mean and standard error of the performances.
We used different initial settings for the test. In the source environment, we did not use the RL environment and just pre-trained the model from demonstration data.
It should be noted that our purpose is not to develop a state-of-the-art MARL algorithms and the strengths of our approach are to enable us to apply it to many existing methods and to provide us with an RL model striking a balance between reproducibility as imitation and generalization.

\begin{figure*}[t]
\centering
\includegraphics[width=1\linewidth]{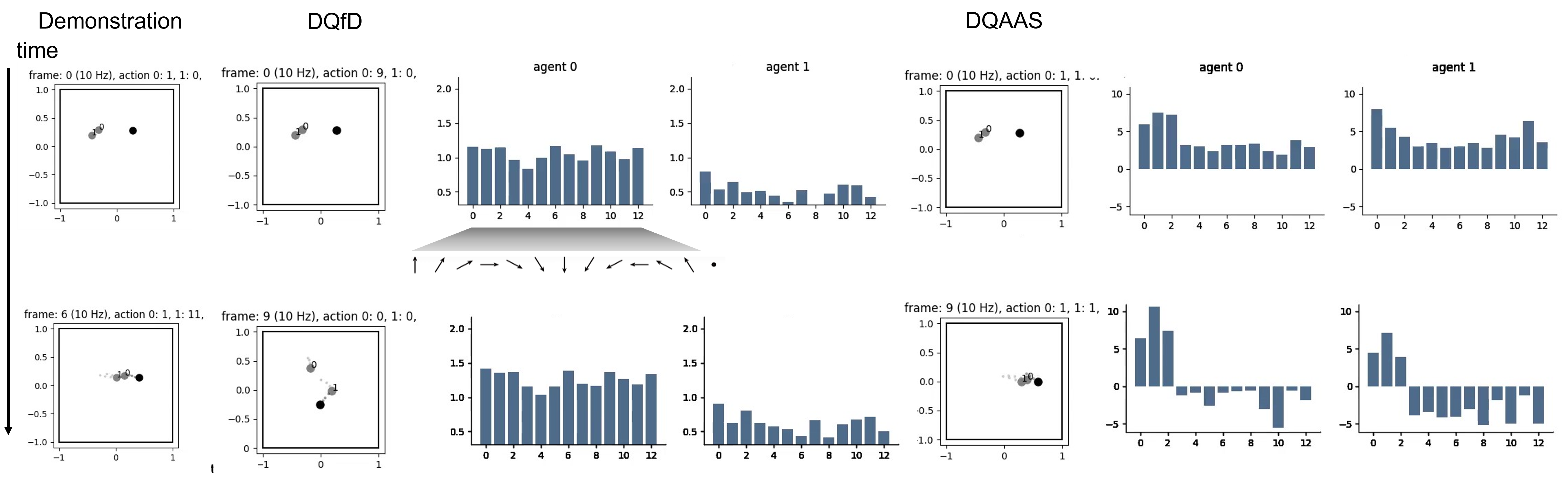}
\vspace{-10pt}
\caption[]{Example RL results of the baseline (DQfD \cite{hester2018deep}, center) and our approach (DQAAS, right), and the demonstration in the source domain (left) in the 2vs1 chase-and-escape task.
Histograms are the Q-function values for each action. 
There are 13 actions including acceleration in 12 directions every 30 degrees in the relative coordinate system (action 0 means moving towards the prey) and doing nothing (action 12: round point).
}
\vspace{-10pt}
\label{fig:ex_chase2vs1}
\end{figure*}

\vspace{-10pt}
\subsection{Performance on Chase-and-escape}
\vspace{-5pt}
First, we verified our method using a chase-and-escape task, in which the predators and prey interacted in a two-dimensional world with continuous space and discrete time. 
The numbers of predators and prey were 2 and 1, respectively. 
We first describe the common setting between the source (demonstration) and target RL tasks. 
The environment was constructed by modifying an environment called the predator-prey in Multi-Agent Particle Environment (MAPE) \cite{lowe2017multi,tsutsui2022emergence,tsutsui2022collaborative}. 
Following \cite{tsutsui2022collaborative}, the play area size was constrained to the range of -1 to 1 on the $x$ and $y$ axes, all agent (predator/prey) disk diameters were set to 0.1, obstacles were eliminated, and predator-to-predator contact was ignored for simplicity. 
The predators were rewarded for capturing the prey ($+1$), namely contacting the disks, and punished for moving out of the area ($-1$), and the prey was punished for being captured by the predator or for moving out of the area ($-1$).

Fig. \ref{fig:example}a shows an example of the chase-and-escape task.  
The time step was 0.1 s and the time limit in each episode was set to 30 s. 
The initial position of each episode was randomly selected from a range of -0.5 to 0.5 on the $x$ and $y$ axes. 
If the predator captured the prey within the time limit, the predator was successful; otherwise, the prey was successful. 
If one side (predators/prey) moved out of the area, the other side (prey/predators) was deemed successful.
There are 13 actions including acceleration in 12 directions at every 30 degrees in relative coordinate system and doing nothing.
For the relative mobility of predators and prey in both environments, to examine the effect of domain adaptation, we set the same mobility of the prey in the source and target RL environments, but for the predators, we set 120\% and 110\% of the prey mobility in the source and target RL environments, respectively.
The predators did not share the rewards for simplicity. 

Before using RL algorithm, we first created the demonstration dataset using the DQN (without data).
After learning 10 million episodes based on the previous setting \cite{tsutsui2022emergence}, we obtained 500 episodes for demonstration with randomized initial conditions (locations).
We split the datasets into 400, 50, and 50 for training, validation, and testing during pre-training, respectively.
Then we pre-trained and trained the models according to Section \ref{sec:proposed}.
To examine the learning performance of the predator movements with the fixed prey movements, we performed RL of all predators with the learned (and fixed) prey. 
For the target RL, we used 50 train and 10 test episodes from the above 100 episodes for pre-train validation and testing (i.e., we did not use the test condition in the target RL during the pre-training and training phases).

The model performance was evaluated by computational simulation of the 10 test episodes as the test phase using the trained models. The termination conditions in each episode were the same as in training. 
We calculated and analyzed the proportion of successful predation and DTW distance between the source and target trajectories in the test phase.

\begin{figure*}[t]
\centering
\includegraphics[width=1\linewidth]{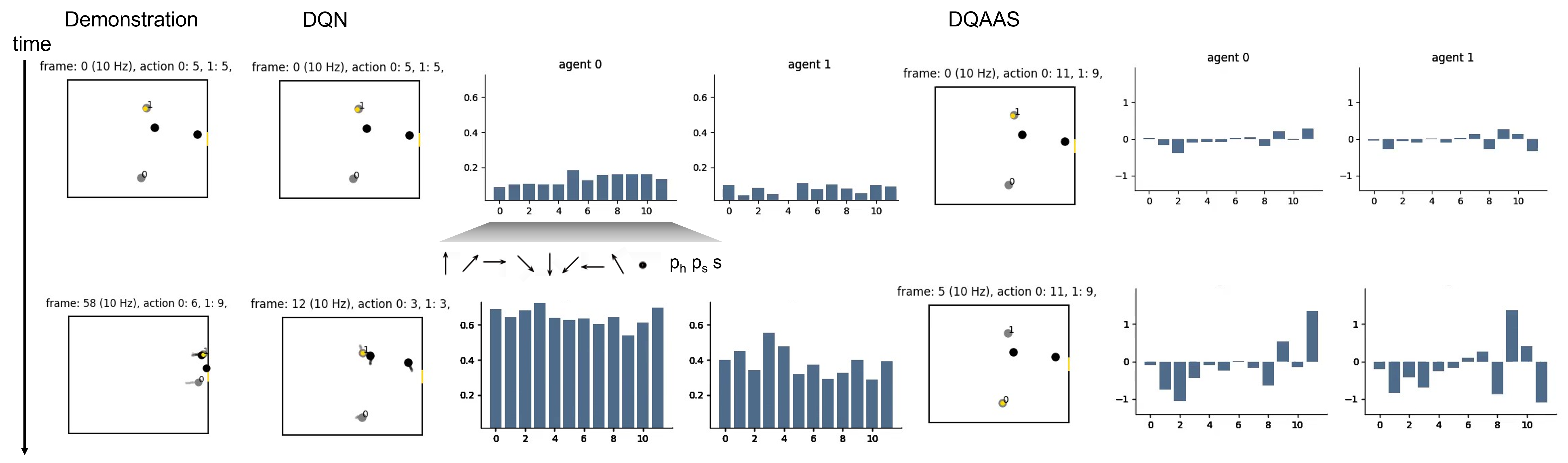}
\vspace{-10pt}
\caption[]{Example RL results of the baseline (DQN, center) and our approach (DQAAS, right), and the demonstration (left) in the 2vs2 football task.
In the demonstration and DQAAS, the agents obtained the goal, but the DQN failed. 
Configurations are the same as Fig. \ref{fig:ex_chase2vs1}.
There are 12 actions including the movement in 8 directions (with constant velocity) every 45 degrees in the relative coordinate system (actions 0-7 and action 0 means moving toward the center of the goal), doing nothing (action 8: round point), and high pass (action 9: $p_h$), short pass (action 10: $p_s$), and shot (action 11: $s$), which are partially based on GFootball \cite{kurach2020google}.
}
\label{fig:ex_nuf2vs2}
\vspace{-10pt}
\end{figure*}

\vspace{-0pt}
\newcommand{\md}[2]{\multicolumn{#1}{c|}{#2}}
\newcommand{\me}[2]{\multicolumn{#1}{c}{#2}}
\begin{table}[ht!]
\centering
\scalebox{0.75}{
\begin{tabular}{l|cc|cc}
\Xhline{3\arrayrulewidth} 
& \md{2}{Reward} & \me{2}{DTW distance} \\ 
& \me{1}{pre-trained} & \md{1}{0.5M steps} &  \me{1}{pre-trained} & \me{1}{0.5M steps} \\
\hline
DQN & 0.04 $\pm$ 0.06 & 0.11 $\pm$ 0.03 & \textbf{4.12} $\pm$ \textbf{0.73} & 5.70 $\pm$ 0.62 \\
\hline
DQfD & 0.00 $\pm$ 0.00 & 0.04 $\pm$ 0.03 & 5.02 $\pm$ 0.34 & 4.94 $\pm$ 0.23 \\
DQfAD & 0.00 $\pm$ 0.00 & 0.06 $\pm$ 0.03 & 5.02 $\pm$ 0.34 & 4.80 $\pm$ 0.57 \\
\hline
DQAS & \textbf{0.25} $\pm$ \textbf{0.08} & \textbf{0.26} $\pm$ \textbf{0.08} & 5.37 $\pm$ 0.40 & 4.97 $\pm$ 1.16 \\
DQAAS & \textbf{0.25} $\pm$ \textbf{0.08} & \textbf{0.29} $\pm$ \textbf{0.09} & 5.37 $\pm$ 0.40 & \textbf{4.73} $\pm$ \textbf{1.07} \\
\Xhline{3\arrayrulewidth} 
\end{tabular}
}
\caption{\label{tab:results_chase2vs1} Performance on 2vs1 chase-and-escape task.}
\vspace{-0pt}
\end{table}

We then show the proportion of successful predation and DTW distance between the source and target trajectories for each model in Table \ref{tab:results_chase2vs1}.
The results show that our approaches (DQAAS) achieved better performances for both indicators than baselines. 
The obtained rewards and DTW distance had a trade-off relationship. In general, how to strike a balance is not obvious.
In this task, with increased training steps, the DQAAS first learned the ability to maximize a reward and then learned the reproducibility at the expense of the reward. 

Here we show example results of the baseline (DQfD \cite{hester2018deep}) and our approach (DQAAS) in Fig \ref{fig:ex_chase2vs1}.
The demonstration (left) shows that two predators chased the prey almost linearly and caught the prey in this scenario. 
In the source domain (demonstration), the predators were much faster than the prey (120 \%), but in the target domain (RL), the predators were only slightly faster than the prey (110 \%).
Then, the task becomes more challenging than in the source domain and learning the Q-function correctly becomes more important to catch prey.
Compared with the baseline, our approach correctly learned Q-function values in which the distribution concentrated near action 0 (here action 0 means moving toward the prey). 
These results imply the effectiveness of our approach quantitatively and qualitatively. 

\vspace{-5pt}
\subsection{Performance on Football Tasks}
\vspace{-5pt}
Next, we used real-world demonstrations of football players and verified our method.
We created an original football environment (called NFootball) in our provided code because in a recent popular environment (GFootball \cite{kurach2020google}) the transition algorithms are difficult to customize and some commands (e.g., pass) did not work well within our intended timings. 
NFootball has a simple football environment and all algorithms are written in Python and then transparent. 
Similarly to GFootball and MAPE environments, players interacted in a two-dimensional world with continuous space and discrete time.
The play pitch size was the same as GFootball \cite{kurach2020google}: a range of -1 to 1 and -0.42 to 0.42 on the $x$ and $y$ axes, respectively, and the goal on the $y$ axis was in the range of -0.044 to 0.044.

\begin{figure*}[t]
\centering
\includegraphics[width=0.8\linewidth]{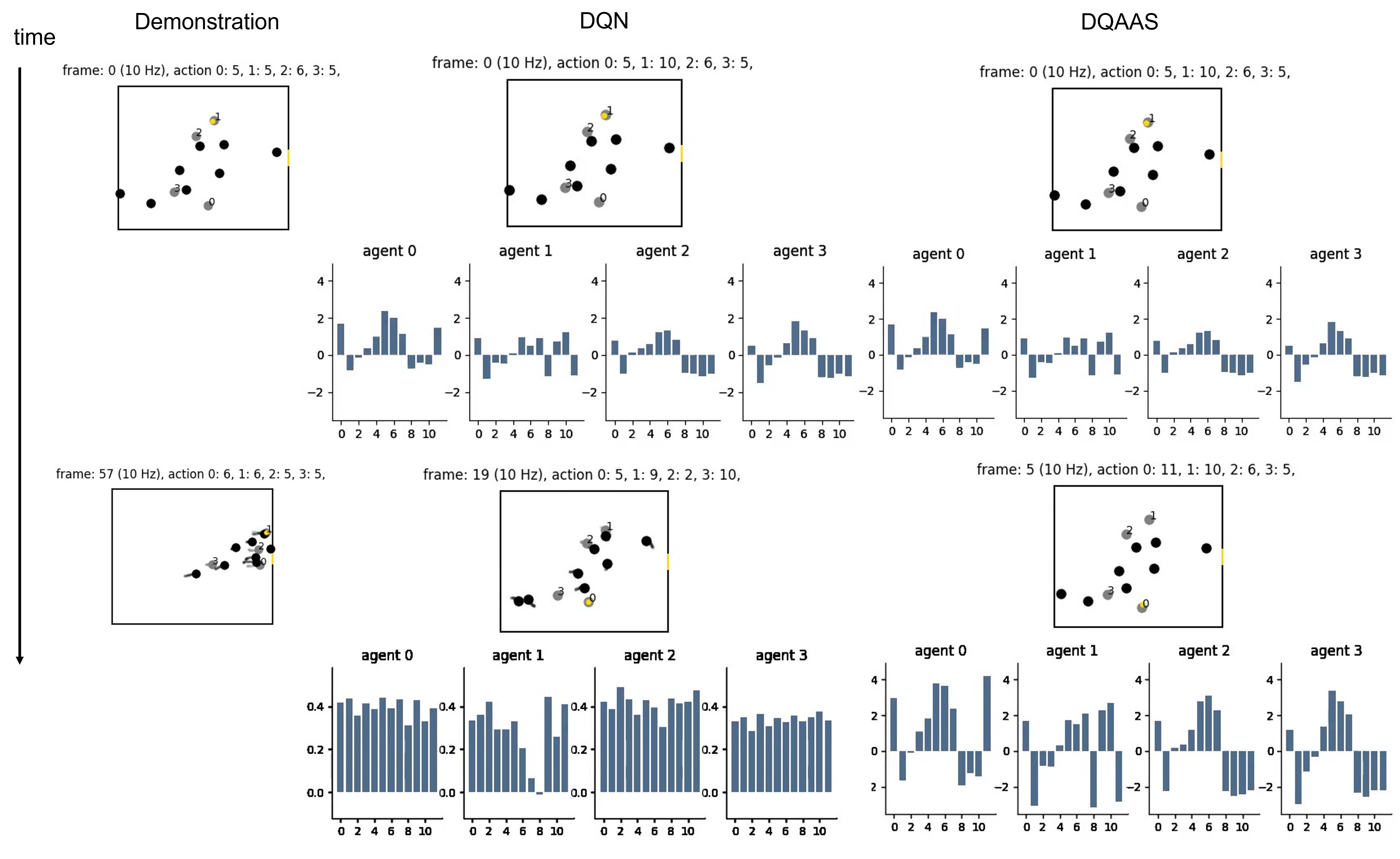}
\caption[]{Example RL results of the baseline (DQN, center) and our approach (DQAAS, right), and the demonstration (left) in the 4vs8 football task.
Configurations are similar to Fig. \ref{fig:ex_nuf2vs2} (in this task, 4 attackers are learned).
In the demonstration and DQAAS, the agents obtained the goal, but the DQN failed. 
The action space is the same as the 2vs2 football task.
}
\label{fig:ex_nuf4vs8}
\end{figure*} 

Figs. \ref{fig:example}b and c show examples of two football tasks: 2vs2 and 4vs8, respectively.
The initial position of each episode was selected as the last passer's possession in the goal scenes based on real-world data as explained below.
The time step was 0.1 s and the time limit in each episode was set to 8.5 s (based on the maximum time length of the real-world data with a margin). 
All attackers and defenders were rewarded and punished for goal ($+10$) and concede ($-10$), respectively.
Also, all defenders and attackers were rewarded and punished for ball gain ($+1$) and lost ($-1$), respectively.
To complete matches, each player is punished for moving out of the pitch ($-5$).
If any reward or punishment is obtained, the episode is finished.
We consider the 2vs2 task for a simple extension of the 2vs1 chase-and-escape task and the 4vs8 task (4 attackers) for more realistic situations. 
Note that currently, the learning of 11vs11 is difficult and time-consuming, and thus we limited the scenarios. 
To examine the learning performances of the attacking movements with the fixed defenders' movements, we first performed the RL of all players with the DQAAS algorithm, and then we performed RL of all attackers with the learned (and fixed) defenders. 
The mobilities of the attackers and defenders are the same. 
There are 12 actions including the movement in 8 directions (with constant velocity) every 45 degrees in the relative coordinate system, doing nothing, and high pass, short pass, and shot, which are partially based on GFootball \cite{kurach2020google}.

Before using RL algorithm, we first created the demonstration dataset using real-world player location data in professional soccer league games.
We used the data of 54 games in the Meiji J1 League 2019 season held in Japan.  
The dataset includes event data (i.e., labels of actions, e.g., passing and shooting, recorded at 30 Hz and the xy coordinates of the ball) and tracking data (i.e., xy coordinates of all players recorded at 25 Hz) provided by Data Stadium Inc.
We extracted 198 last-pass-and-goal sequences and 1,385 last-pass sequences (including a ball lost) for training and pre-training of the RL model from the dataset.
In pre-training, we split the dataset into 1,121 training, 125 validation, and 139 test sequences (or episodes). 
We set shot rewards ($+1$) for the attacker in addition to the above rewards and punishments (but we did not use the out-of-pitch punishment) because the goal reward was sparse and limited.    
In the target RL, we used 16 train and 5 test episodes from the above 198 episodes (we did not use the test condition in the target RL during the pre-training and training phases).
We calculated and analyzed the obtained reward and DTW distance between the source and target trajectories in the test phase.

\vspace{-0pt}
\begin{table}[ht!]
\centering
\scalebox{0.75}{
\begin{tabular}{l|cc|cc}
\Xhline{3\arrayrulewidth} 
& \md{2}{Reward} & \me{2}{DTW distance} \\ 
& \me{1}{pre-trained} & \md{1}{0.5M steps} &  \me{1}{pre-trained} & \me{1}{0.5M steps} \\
\hline
DQN & 0.00 $\pm$ 0.00 & 1.40 $\pm$ 1.16 & 3.15 $\pm$ 0.47 & 2.69 $\pm$ 0.53 \\
\hline
DQfD & 0.00 $\pm$ 0.00 & 0.00 $\pm$ 0.00 & 4.58 $\pm$ 0.00 & 5.18 $\pm$ 0.00 \\
DQfAD & 0.00 $\pm$ 0.00 & 0.00 $\pm$ 0.00 & 4.58 $\pm$ 0.00 & 5.29 $\pm$ 0.01 \\
\hline
DQAS & \textbf{8.00} $\pm$ \textbf{0.00} & \textbf{8.00 }$\pm$ \textbf{0.00} & \textbf{2.25} $\pm$ \textbf{0.00} & \textbf{2.25} $\pm$ \textbf{0.00} \\
DQAAS & \textbf{8.00} $\pm$ \textbf{0.00} & \textbf{8.00} $\pm$ \textbf{0.00} & \textbf{2.25} $\pm$ \textbf{0.00} & \textbf{2.25} $\pm$ \textbf{0.00} \\
\Xhline{3\arrayrulewidth} 
\end{tabular}
}
\caption{\label{tab:results_nuf2vs2} Performance on 2vs2 football task.}
\vspace{-5pt}
\end{table}

Next, we show the quantitative and qualitative performances in the 2vs2 and 4vs8 tasks in this order. 
First, we show the average return and the DTW distance between the source and target trajectories for each model of the 2vs2 task in Table \ref{tab:results_nuf2vs2}. 
The results show that our approaches (DQAS and DQAAS) achieved better performances for both indicators than baselines with demonstrations (DQfD and DQfAD).
However, the two performance indicators in the learning models from demonstrations (except for DQN) did not change according to the learning steps.  
It suggests that the pre-trained models from demonstrations obtained other local solutions and may be struggle to obtain better solutions (in particular, in terms of reproducibility). 
For example, in Fig \ref{fig:ex_nuf2vs2}, we show the demonstration, example results of the baseline without demonstration (DQN), and our approach (DQAAS).
The demonstration (left) shows that the attacker \#1 passed the ball to the attacker \#0  during moving toward the goal.
However, in our approach (right), agents learned movements simply to pass the ball and shoot without moving toward the goal. 
In contrast, the model without demonstration (center) learned moving toward the goal without passing and shooting the ball.
Ideally, combining both generalization and reproducibility will be expected but the domain-specific modeling and reality of the simulator is left for future work in this task. 
In terms of the Q-learning, as shown in Fig. \ref{fig:ex_nuf2vs2}, the agents obtained the goal in the demonstration (left) and DQAAS (right), but the DQN (center) failed. 
Although our approach did not reproduce the demonstration movements toward the goal, compared with DQN, our approach correctly learned Q-function values in which the higher values were observed in actions 10 and 11 for the passer (agent 1) and shooter (agent 0), respectively.

\vspace{-0pt}
\begin{table}[ht!]
\centering
\scalebox{0.70}{
\begin{tabular}{l|cc|cc}
\Xhline{3\arrayrulewidth} 
& \md{2}{Reward} & \me{2}{DTW distance} \\ 
& \me{1}{pre-trained} & \md{1}{0.5M steps} &  \me{1}{pre-trained} & \me{1}{0.5M steps} \\
\hline
\hline
DQN & 0.00 $\pm$ 0.00 & 0.16 $\pm$ 0.13 & \textbf{3.22} $\pm$ \textbf{0.22} & \textbf{3.24} $\pm$ \textbf{0.22 }\\
CDS & 0.12 $\pm$ 0.24 & 0.12 $\pm$ 0.24 & \textbf{3.10} $\pm$ \textbf{0.07} &\textbf{ 3.25} $\pm$\textbf{ 0.12} \\
\hline
DQfD & 0.00 $\pm$ 0.00 & 0.00 $\pm$ 0.00 & 3.35 $\pm$ 0.00 & 3.78 $\pm$ 0.00 \\
DQfAD & 0.00 $\pm$ 0.00 & 0.00 $\pm$ 0.00 & 3.35 $\pm$ 0.00 & 4.28 $\pm$ 0.00 \\
CDS+fD & 0.00 $\pm$ 0.00 & 0.00 $\pm$ 0.00 & 3.71 $\pm$ 0.00 & 3.76 $\pm$ 0.00 \\
CDS+fAD & 0.00 $\pm$ 0.00 & 0.00 $\pm$ 0.00 & 3.71 $\pm$ 0.00 & 3.76 $\pm$ 0.00 \\
\hline
DQAS & 0.00 $\pm$ 0.00 & \textbf{6.00} $\pm$ \textbf{0.00} & 4.54 $\pm$ 0.00 & \textbf{3.27} $\pm$ \textbf{0.00} \\
DQAAS & 0.00 $\pm$ 0.00 & \textbf{6.00} $\pm$ \textbf{0.00} & 4.54 $\pm$ 0.00 & \textbf{3.30} $\pm$ \textbf{0.00} \\
CDS+AS & \textbf{6.00 }$\pm$ \textbf{0.00} & \textbf{6.00 }$\pm$ \textbf{0.00} & \textbf{3.25} $\pm$ \textbf{0.00} & \textbf{3.30} $\pm$ \textbf{0.00} \\
CDS+AAS & \textbf{6.00} $\pm$ \textbf{0.00} & \textbf{6.00} $\pm$ \textbf{0.00 }& \textbf{3.25} $\pm$ \textbf{0.00} & \textbf{3.30} $\pm$ \textbf{0.00} \\
\Xhline{3\arrayrulewidth} 
\end{tabular}
}
\caption{\label{tab:results_nuf4vs8} Performance on 4vs8 football task.}
\vspace{-5pt}
\end{table}

Next, we show the results of the 4vs8 football task.
The quantitative results in Tables \ref{tab:results_nuf4vs8} in DQN-based RL models show that our approaches (DQAS and DQAAS) achieved better performances for both indicators than baselines with demonstrations (DQfD and DQAAS).
These results and discussions were similar to those in the 2vs2 task shown in Table \ref{tab:results_nuf2vs2}. 
In addition, we examined the centralized learning approach using CDS \cite{li2021celebrating}. 
These results shown in Table \ref{tab:results_nuf4vs8} in CDS-based RL models were very similar to those in Table \ref{tab:results_nuf4vs8} in DQN-based RL models. 
We confirmed that the cause of the reproducibility issue may not be the centralized/decentralized or classic/recent deep RL. 
More task-specific modeling using domain knowledge \cite{zare2021improving,nguyen2020structure} can be a possible solution, which is left for future work.
In terms of Q-learning (Fig. \ref{fig:ex_nuf4vs8}), compared with DQN (center), our approach (rights) correctly learned Q-function values for actions 10 and 11 for the passer and shooter, which were similar results to those in Fig. \ref{fig:ex_nuf2vs2}. 
If the model can imitate behaviors of players in the real-world football, we can estimate values for their behaviors and decision making using estimated Q-function values, which may be difficult for either data-driven and RL approaches. 

\vspace{-15pt}
\section{\uppercase{Conclusion}}
\label{sec:conclusion}
\vspace{-5pt}
We proposed a novel method for domain adaptation in RL from real-world multi-agent demonstration, which will bridge the gap between RL in cyberspace and data-driven modeling. 
In the experiments, using chase-and-escape and football tasks with the different dynamics between the unknown source and target environments, we showed that our approach balanced between the reproducibility and generalization more effectively than the baselines. In particular, we used the tracking data of professional football players as expert demonstrations in a football RL environment and demonstrated successful performances in both despite the larger gap between behaviors in the source and target environments than in the chase-and-escape task.
Possible future research directions are to create a better multi-agent simulator and RL model utilizing domain knowledge for reproducing not only actions but also movements such as used by \cite{tsutsui2023synergizing}.
In another direction, although modeling football movements would be currently challenging, for example, application to multi-animal behaviors will provide more scientifically valuable insights.


\ifarxiv
\vspace{-15pt}
\section*{\uppercase{Acknowledgements}}
This work was supported by JSPS KAKENHI (Grant Numbers 21H04892, 21H05300 and 23H03282) and JST PRESTO (JPMJPR20CA).
\fi

\vspace{-15pt}
\bibliographystyle{apalike}
{\small
\bibliography{main}}

\end{document}